\documentclass{article}

\usepackage{PRIMEarxiv}

\usepackage[utf8]{inputenc} % allow utf-8 input
\usepackage[T1]{fontenc}    % use 8-bit T1 fonts
\usepackage{hyperref}       % hyperlinks
\usepackage{url}            % simple URL typesetting
\usepackage{booktabs}       % professional-quality tables
\usepackage{amsfonts}       % blackboard math symbols
\usepackage{nicefrac}       % compact symbols for 1/2, etc.
\usepackage{microtype}      % microtypography
\usepackage{lipsum}
\usepackage{fancyhdr}       % header
\usepackage{graphicx}       % graphics
\graphicspath{{media/}}     % organize your images and other figures under media/ folder

\usepackage{xcolor}
\usepackage{amsmath,amssymb,amsfonts}
\usepackage{multirow}
\usepackage{float}

\title{PedsFM: Learning Pediatric Trajectories at Native Physiological Resolution with a Hybrid Discrete-Continuous Architecture}

\author{
\parbox{\textwidth}{
\centering
\normalfont\mdseries\small
Daniela Chanci\textsuperscript{1*},
Baraa Abed\textsuperscript{2},
Mehak Arora\textsuperscript{2},
Henry P. Foote\textsuperscript{3},
Kush Narang\textsuperscript{4},
Noah Stanco\textsuperscript{3},
Tasnuba Islam\textsuperscript{2},
Alexandre T. Rotta\textsuperscript{3},
Christoph P. Hornik\textsuperscript{3},
Ioannis Koutroulis\textsuperscript{5},
Jennifer A. Muszynski\textsuperscript{6},
Mark S. Wainwright\textsuperscript{7},
Akira Nishisaki\textsuperscript{8},
Sanjiv D. Meta\textsuperscript{8},
Neel Shah\textsuperscript{9},
Daniel Tawfik\textsuperscript{10},
Shu-Ling Chong\textsuperscript{11},
Jan Hau Lee\textsuperscript{12},
Aaron Kornblith\textsuperscript{4},
Chris A. Rees\textsuperscript{13},
Stephanie Brown\textsuperscript{14},
Rishikesan Kamaleswaran\textsuperscript{15,16*}
}
\\[1.5em]
\parbox{0.95\textwidth}{
\centering
\normalfont\mdseries\footnotesize
\textsuperscript{1}Department of Biomedical Engineering, Duke University \\
\textsuperscript{2}Department of Electrical and Computer Engineering, Duke University \\
\textsuperscript{3}Department of Pediatrics, Duke University School of Medicine \\
\textsuperscript{4}Department of Emergency Medicine, University of California San Francisco \\
\textsuperscript{5}Department of Pediatrics, George Washington University, Children's National Hospital \\
\textsuperscript{6}Department of Pediatrics, The Ohio State University College of Medicine, Nationwide Children's Hospital \\
\textsuperscript{7}Department of Pediatrics, The University of Washington, Seattle Children's Hospital \\
\textsuperscript{8}Department of Anesthesiology and Critical Care Medicine, Children's Hospital of Philadelphia \\
\textsuperscript{9}Department of Pediatrics, Washington University in St. Louis \\
\textsuperscript{10}Division of Pediatric Critical Care Medicine, Department of Pediatrics, Stanford University School of Medicine \\
\textsuperscript{11}Department of Emergency Medicine, KK Women's and Children's Hospital \\
\textsuperscript{12}Children's Intensive Care Unit, KK Women's and Children's Hospital \\
\textsuperscript{13}Division of Pediatric Emergency Medicine, Emory University School of Medicine, Children's Healthcare of Atlanta \\
\textsuperscript{14}Department of Pediatrics, Emory University School of Medicine, Children's Healthcare of Atlanta \\
\textsuperscript{15}Department of Surgery, Duke University School of Medicine \\
\textsuperscript{16}Department of Anesthesiology, Duke University School of Medicine \\[0.6em]
\textsuperscript{*}Corresponding authors:
\texttt{daniela.chanciarrubla@duke.edu, r.kamaleswaran@duke.edu}
}
}

\begin{document}

\maketitle

\begin{abstract}

Electronic health record (EHR) foundation models, leveraging their self-supervised training objectives, offer a path to clinically actionable insight without task-specific labeling. However, existing approaches are developed and validated primarily in adult populations, often discretize continuous physiological measurements, and exclude dense signals such as vital signs. These limitations are particularly relevant in pediatrics, where interpretation of clinical numerical variables is strongly age-dependent. Additionally, children remain severely underrepresented in healthcare machine learning research. We introduce PedsFM, a pediatric-specific, decoder-only generative foundation model that jointly represents discrete clinical events and continuous physiological measurements at their native numerical resolution through specific embedding pathways and dual discrete/continuous prediction heads. This model was pretrained on 65,228 hospitalizations involving at least one pediatric intensive care unit admission across three U.S. hospitals, comprising 237 million training tokens. We evaluated the model along three complementary axes: generative plausibility, zero-shot probabilistic prediction of clinically relevant outcomes, and representation learning via linear probing. Generated vital sign trajectories were physiologically plausible and outperformed a last-seen-value carry-forward baseline for most vitals, and conditioning generation on medications produced pharmacologically coherent shifts in predicted physiology. Zero-shot predictions outperformed a fully supervised, task-specific XGBoost baseline for four laboratory abnormality tasks, while mean-pooled representations achieved comparable performance for early in-hospital mortality and vasopressor initiation, and outperformed last-token representations across all six downstream tasks. These findings suggest that resolution-preserving, hybrid generative modeling can extract clinically meaningful insights from pediatric critical care trajectories without task-specific training, and motivate further development of pediatric-specific foundation models.

\end{abstract}

\section{Introduction}
\label{sec:intro}

Electronic Health Records (EHR) data capture the clinical trajectory of patients through longitudinal recordings of vital signs, laboratory tests, medications, procedures, and other clinical events \cite{xiao_opportunities_2018}. These data present an ideal opportunity to develop data-driven decision support tools that can generate actionable insights from routinely collected clinical information. Consequently, over the last decade, numerous studies have focused on developing machine learning methods for a wide variety of healthcare applications \cite{shickel_deep_2018}. Traditional machine learning approaches usually rely on carefully engineered features, task-specific cohorts, and expert-defined labels, and they have demonstrated strong discriminative performance for well-specified clinical outcomes. However, task-specific models are developed for a particular prediction task, requiring the corresponding cohort, outcome labels, feature extraction, and model training procedure to be defined for each new outcome \cite{moor_foundation_2023, bommasani_opportunities_2021}.

Advances in sequence modeling, such as the transformer architecture, have provided suitable methods for representing longitudinal, structured EHR data \cite{vaswani_attention_2017}. Additionally, transformer-based foundation models offer an alternative to training a separate model for each task and have become central to recent progress in artificial intelligence in healthcare \cite{guo_systematic_2026}.  These models are pretrained on large datasets using self-supervised objectives and can be subsequently adapted to a variety of downstream applications,  which may reduce the number of labeled examples required for any individual task \cite{wornow_ehrshot_2023}. In structured EHR modeling, patient records are commonly represented as sequences of timestamped clinical events, allowing transformer-based models to learn contextual representations of longitudinal clinical trajectories that can be reused across tasks. Early EHR foundation models, such as BEHRT and Med-BERT, leveraged bidirectional transformer architectures to learn complex patient representations through a self-supervised objective, improving performance across downstream prediction tasks \cite{li_behrt_2020, rasmy_med-bert_2021}.

More recently, decoder-only architectures with generative capabilities have emerged as a promising approach for EHR modeling \cite{hao_gpt-based_2024}. These models learn the conditional distribution of future clinical events given a prior patient history, allowing autoregressive generation of patient trajectories. Models such as ETHOS and Curiosity have demonstrated that forecasted patient trajectories can support zero-shot clinical predictions without task-specific training or fine-tuning \cite{renc_zero_2024, waxler_generative_2025}. Similarly, CLMBR-T, released with the EHRSHOT benchmark, demonstrated that pretrained patient representations can improve sample efficiency across different downstream clinical prediction tasks \cite{wornow_ehrshot_2023}. However, most existing EHR foundation models rely primarily on demographics, laboratory orders, diagnoses, medications, and procedures, while excluding dense physiological signals, such as vital signs, that convey valuable insight into a patient's state. Also, numerical values are often discretized into categorical tokens using population-level bins, which can facilitate their inclusion into the autoregressive framework, but it can obscure clinically meaningful variation \cite{renc_zero_2024, waxler_generative_2025, wornow_shaky_2023}. It has been demonstrated that the inclusion of numerical values in modeling EHR trajectories can improve the prediction of subsequent clinical events \cite{draxler_transformers_2025}. Besides, discretizing such values can be particularly limiting in pediatric patients, where the interpretation of a given value is strongly age-dependent \cite{dufendach_special_2024}. 

Children are severely underrepresented in machine learning research in healthcare \cite{richter_toward_2025}. However, pediatrics is an area in which foundation models could offer substantial benefit, since labeled pediatric datasets are typically smaller and clinically relevant outcomes are rare \cite{muralidharan_recommendations_2023, hua_underrepresentation_2026}. Additionally, there are significant differences between children and adults in physiology, disease prevalence, treatment response, and expected ranges for common clinical measurements, such as heart rate and respiratory rate \cite{fleming_normal_2011}. These differences limit the applicability of models developed primarily on adult data, which have been shown to degrade when applied to children \cite{yeo_pediatric_2026}. Hence, a model trained directly on pediatric trajectories may better capture patterns of physiology and care that are specific to children \cite{muralidharan_recommendations_2023}. However, pediatric-specific EHR foundation models remain largely unexplored. A systematic review on foundation models for structured EHR published recently found that out of the 53 models available in the literature as of 2025, none of them was pretrained specifically on pediatric data \cite{guo_systematic_2026}. Earlier this year, a decoder-only pediatric foundation model, TEDDY, was developed using EHR data from a single U.S. pediatric institution \cite{neeley_teddy_2026}. This model largely focused on longitudinal diagnostic codes, whereas pediatric critical care is characterized by rapidly evolving physiological measurements, laboratory values, medications, and interventions that are not captured by diagnosis codes alone, limiting its ability to capture subtle changes in patient state and prioritizing long-horizon new diagnoses predictions.

In this work, we introduce PedsFM, a generative foundation model for structured, longitudinal pediatric EHR data that enables the joint prediction of discrete clinical events and their associated continuous values. PedsFM is a decoder-only transformer pretrained on millions of clinical events obtained from three hospitals within two U.S. healthcare systems, yielding dense trajectories for each hospitalization involving at least one PICU admission. We evaluate the model using autoregressive generations for zero-shot predictions of clinically relevant outcomes, including early in-hospital mortality, early vasopressors initiation, and laboratory abnormalities, without additional task-specific training, along with assessing the learned representations. Our findings demonstrate that pediatric-specific generative pretraining can support clinically meaningful zero-shot prediction from dense critical care trajectories while learning transferable patient representations, highlighting the complementary value of direct probabilistic inference and trajectory-level representation learning for pediatric clinical applications.

\section{Methods}
\label{sec:methods}

\subsection{Study Design and Data Preprocessing}
\label{sec:dataset}

We conducted a retrospective, multicenter study using longitudinal EHR data from three hospitals that provide specialized pediatric care in the United States. The study population consisted of pediatric hospitalizations, patients \(<\) 18 years old at admission, occurring between 2010 and 2024, that included at least one admission to the PICU. For each hospitalization, we constructed a clinical trajectory from the start of the encounter to discharge. We used chronologically ordered clinical events, including demographics, encounter events, unit transfers, vital signs, laboratory tests, medications, body measurements, imaging procedures, respiratory procedures (e.g., endotracheal intubation and tracheostomy), and mechanical ventilation settings. It is important to note that even though diagnoses are present in the EHR data in the form of ICD-10 codes, they were not included in this study since they can be assigned and revised after the clinical events they represent, and are generally finalized after discharge. Therefore, the inclusion of ICD-10 codes could introduce data leakage when modeling clinical trajectories \cite{ramadan_diagnostic_2025}.

The dataset was partitioned into training, validation, and test splits (85/5/10) at the patient level, therefore, all hospital encounters from a single patient were assigned to the same split to avoid data leakage. We applied a minimal preprocessing pipeline consisting of three steps. The first step was to harmonize the data, unifying variable names and variable units across institutions. The second step consisted of removing outliers for numerical events by discarding data below the 0.5$^{th}$ percentile and above the 99.5$^{th}$ percentile. The last step was the normalization of numerical events using z-score standardization. Before normalization, some predefined variables were transformed using the natural logarithm, i.e., $log(x+1)$. Also, before normalized numerical values were provided as input to the model, they were additionally clipped to the $[-10, 10]$ interval to prevent extreme values that remained after outlier removal and improve training stability. Outlier bounds and normalization parameters were computed exclusively using the training split. 

This study was approved by the Institutional Review Board of Duke University (Pro00116916).

\subsection{Tokenization Strategy}
\label{sec:tokenization}

We represented each hospitalization as a sequence of interleaved discrete tokens and
continuous numerical values. The trajectories begin with a fixed block of five demographic tokens (age group, age in years, ethnicity, sex, and race), followed by chronologically ordered clinical events. The overall tokenization strategy is summarized in Supplementary Table \ref{tab:tokens}.

\paragraph*{Discrete Clinical Events.} Discrete clinical events include demographics, encounter events, unit transfers, laboratory specimen collections, medications, imaging procedures, and respiratory procedures. They were represented by a single token or multiple tokens with a hierarchical structure to keep the vocabulary size small. For example, hospital discharge tokens were always followed by a discharge disposition token. Similarly, imaging procedures were represented by an imaging modality token followed by the corresponding anatomic location token. Finally, medications, which used the Anatomical Therapeutic Chemical (ATC) codes inspired by \cite{renc_zero_2024}, were the clinical event represented by the largest number of tokens in the following order: medication administration action, administration route, ATC anatomical main group, ATC therapeutic subgroup, ATC pharmacological subgroup, and ATC chemical component if present.

\paragraph*{Continuous Clinical Events.} Continuous clinical events include vital signs, laboratory results, body measurements, and mechanical ventilation settings. Instead of discretizing the associated numerical values, we represented them as continuous normalized scalars to avoid the loss of clinically meaningful resolution. This is particularly important in pediatrics given the high variability in normal ranges among clinical characteristics across different age groups. As a result, each of the continuous clinical events was represented by two consecutive elements, a discrete token indicating the event type and the corresponding variable name, followed by the normalized continuous value. 

\paragraph*{Temporal Tokens.} There were two different types of temporal tokens: start of the sequence and delta tokens. The first was added after the demographic tokens to indicate the start of the chronologically ordered clinical sequence. Therefore, it appeared only one time per trajectory. The second was assigned when the elapsed time between two consecutive clinical events exceeded five minutes, adapted from \cite{renc_zero_2024}. Thus, the appropriate delta token appeared multiple times throughout the trajectory.

\subsection{Model Architecture}
\label{sec:architecture}

\begin{figure}[h]
\begin{center}
\includegraphics[width=1.0\linewidth]{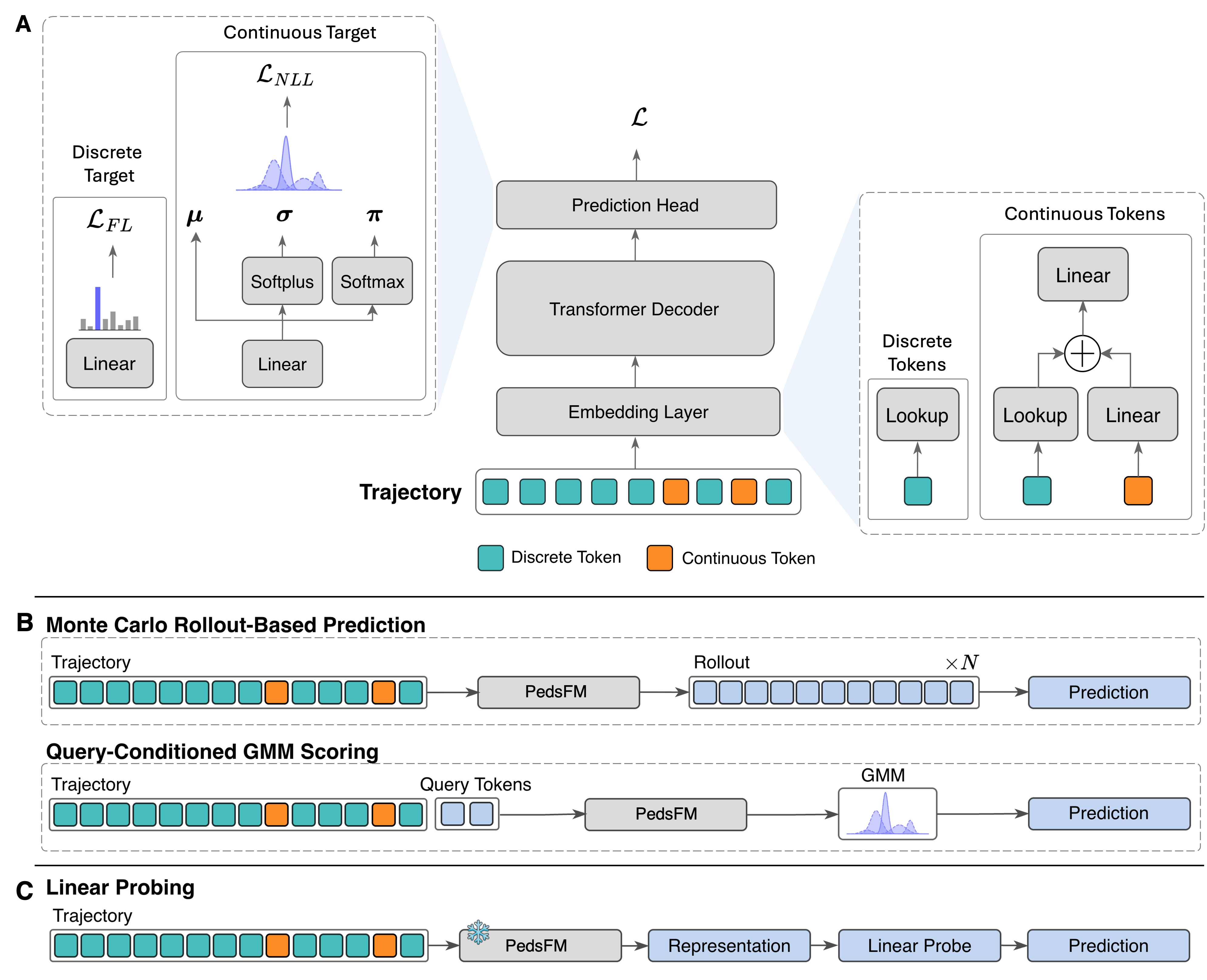}
\end{center}
\caption{PedsFM architecture and evaluation. A) Hybrid architecture with different embedding pathways for discrete/continuous input tokens and dual predictions heads for discrete/continuous target tokens. B) Zero-shot evaluation strategies: Monte Carlo rollout-based prediction and query-conditioned GMM scoring. C) Linear probing for representation learning evaluation on downstream tasks.}
\label{fig:pedsfm_architecture}
\end{figure}

PedsFM leverages the decoder-only transformer architecture to model the conditional distribution of each clinical event given all preceding events in the hospitalization. We used a deterministic routing mechanism to model interleaved discrete and continuous events (see Figure \ref{fig:pedsfm_architecture}). Hybrid autoregressive architectures have been studied for multimodal generation \cite{tschannen_jetformer_2025} and mixed-type event sequence modeling \cite{draxler_transformers_2025}. PedsFM adapts this dual discrete/continuous prediction heads framework within a single autoregressive sequence for heterogeneous pediatric EHR trajectories, using a language-modeling head for discrete events and a Gaussian mixture model (GMM) head for continuous physiological measurements \cite{tschannen_jetformer_2025}.

\paragraph*{Problem Formulation.} For a hospitalization, let $\mathbf{x}=(x_1,x_2,\ldots,x_T)$
denote a chronologically ordered EHR trajectory obtained after applying the proposed tokenization strategy. Each element of the sequence is associated with a routing state $r_i\in\{0,1,2\}$, where $r_i=0$ denotes a regular discrete token not associated with a numerical value, $r_i=1$ denotes a discrete token that is immediately followed by its associated numerical value, and $r_i=2$ denotes the numerical value itself. Let $\mathcal V$ denote the discrete
vocabulary. The domain of each trajectory element is determined by the routing state:

\[
x_i
\in
\begin{cases}
\mathcal V, & r_i\in\{0,1\},\\[4pt]
\mathbb R, & r_i=2.
\end{cases}
\]

Therefore, continuous numerical values are elements of the trajectory but are not included in the discrete token vocabulary $\mathcal V$. PedsFM is trained using next-token prediction to model each element of the trajectory using preceding elements $P(x_1, \dots, x_T) = \prod_{i=1}^{T} P(x_i \mid x_{<i})$.

\paragraph*{Input Embedding Pathways.} Let $d$ denote the model dimension and $\mathbf{E}\in\mathbb{R}^{|\mathcal V|\times d}$ the learned embedding matrix for the discrete vocabulary. For a discrete token $x_i$, the input representation is obtained through a learned embedding lookup:

\[
    \mathbf{e}_i^{\mathrm{disc}}
    =
    \mathbf{E}[x_i],
    \qquad
    \mathbf{e}_i^{\mathrm{disc}}\in\mathbb{R}^{d}
\]

where $\mathbf{E}[x_i]$ denotes its corresponding learned embedding vector. 

For a continuous value $x_i$, the value is first projected into the model dimension:

\[
    \mathbf{v}_i
    =
    \mathbf{W}_{\mathrm{val}} x_i
    +
    \mathbf{b}_{\mathrm{val}},
    \qquad
    \mathbf{v}_i \in \mathbb{R}^{d}
\]

where $\mathbf{W}_{\mathrm{val}}\in\mathbb R^{d\times1}$ is a learned projection matrix, and $\mathbf{b}_{\mathrm{val}}\in\mathbb R^d$ is a learned bias vector. This
transformation maps the numerical value into the model dimension space.
The resulting vector is then concatenated with the embedding of the immediately
preceding discrete token, $\mathbf{e}_{i-1}^{\mathrm{disc}}$, and projected
again into the model dimension:

\[
    \mathbf{e}_i^{\mathrm{num}}
    =
    \mathbf{W}_{\mathrm{concat}}
    \left[
        \mathbf{v}_i
        \,\Vert\,
        \mathbf{e}_{i-1}^{\mathrm{disc}}
    \right],
    \qquad
    \mathbf{e}_i^{\mathrm{num}}\in\mathbb R^d
\]

where $\Vert$ denotes concatenation and $\mathbf{W}_{\mathrm{concat}} \in \mathbb{R}^{d \times 2d}$. The concatenated vector belongs to $\mathbb R^{2d}$, and
$\mathbf{W}_{\mathrm{concat}}$ projects it back into the model dimension.

The final interleaved input embedding is therefore a sequence of $\mathbf{e}_i$, where

\[
    \mathbf{e}_i =
    \begin{cases}
        \mathbf{e}_i^{\mathrm{disc}},
        & \text{if } x_i \text{ is discrete}, \\[4pt]
        \mathbf{e}_i^{\mathrm{num}},
        & \text{if } x_i \text{ is continuous}
    \end{cases}
\]

The inclusion of the embedding of the immediately preceding token into the calculation of the embedding of the corresponding numerical value allows the model to interpret the same normalized value differently depending on the associate clinical variable. Therefore, the same numerical value will have a different representation for heart rate than for blood pressure or a laboratory result.

\paragraph*{Transformer Backbone.} The sequence of interleaved embeddings is processed by a decoder-only transformer that consists of layers with pre-layer-normalization, multi-head causal self-attention, and a GELU-activated feed forward network (FFN). 

Let $f_{\mathrm{T}}$ denote the transformer backbone. The contextualized representation at each position $i$ of the trajectory is given by

\[
    \mathbf{h}_i
    =
    f_{\mathrm{T}}
    \left(
        \mathbf{e}_{\leq i}
    \right),
    \qquad
    \mathbf{h}_i \in \mathbb{R}^{d},
\]

where $\mathbf{e}_{\leq i}=(\mathbf{e}_1,\ldots,\mathbf{e}_i)$ represents the sequence of input embeddings up to position $i$. This causal structure of the model, in which $\mathbf{h}_i$ depends only on the preceding trajectory is achieved via the causal attention mask. Additionally, positional information is injected via Rotary Position Embeddings (RoPE) within the attention mechanism.

\paragraph*{Dual Prediction Heads.}
The model uses two complementary output heads, a language modeling head and a GMM head, depending on whether the target is discrete or continuous, respectively. Let $\mathbf{h}_i \in \mathbb{R}^{d}$ denote the representation produced by the transformer at position $i$.

For a discrete target, the language modeling (LM) head projects the hidden representation to logits over the discrete vocabulary:

\[
    \mathbf{z}_i
    =
    \mathbf{W}_{\mathrm{LM}}\mathbf{h}_i,
    \qquad
    \mathbf{z}_i\in\mathbb{R}^{|\mathcal V|},
\]

where $\mathbf{W}_{\mathrm{LM}}\in\mathbb{R}^{|\mathcal V|\times d}$ is the output projection matrix. The corresponding categorical distribution is

\[
    \mathbf{q}_i
    =
    \operatorname{softmax}(\mathbf{z}_i),
    \qquad
    \mathbf{q}_i\in\mathbb{R}^{|\mathcal V|},
\]

where $q_{i,v}$ represents the predicted probability of discrete token $v\in\mathcal V$. 

For a continuous target, the GMM head maps the hidden representation to the parameters of a Gaussian mixture distribution with $K$ components. Therefore, there are $3K$ outputs in total

\[
    \begin{bmatrix}
        \boldsymbol{\mu}_i\\
        \widetilde{\boldsymbol{\sigma}}_i\\
        \boldsymbol{\alpha}_i
    \end{bmatrix}
    =
    \mathbf{W}_{\mathrm{GMM}}\mathbf{h}_i
    +
    \mathbf{b}_{\mathrm{GMM}},
\]

where $\mathbf{W}_{\mathrm{GMM}}\in\mathbb{R}^{3K\times d}$ and $\mathbf{b}_{\mathrm{GMM}}\in\mathbb{R}^{3K}$ are learned parameters and $\boldsymbol{\mu}_i,\widetilde{\boldsymbol{\sigma}}_i, \boldsymbol{\alpha}_i\in\mathbb{R}^{K}$.
The component standard deviations and mixture weights are then obtained as

\[
    \boldsymbol{\sigma}_i
    =
    \operatorname{softplus}
    \left(
        \widetilde{\boldsymbol{\sigma}}_i
    \right)
    +
    \epsilon,
\]
\[
    \boldsymbol{\pi}_i
    =
    \operatorname{softmax}
    \left(
        \boldsymbol{\alpha}_i
    \right).
\]

The predictive distribution for the next element $x_{i+1}$ is selected deterministically according to the routing state of the current element $x_{i}$, as explained above. When $r_i=1$, $x_i$ is a discrete token associated with a numerical value, therefore, $x_{i+1}\in\mathbb{R}$ is obtained through the GMM head. Otherwise, if $r_i\in\{0,2\}$, the target is a discrete token obtained through the language modeling head:

\[
    p(x_{i+1}\mid x_{\leq i})
    =
    \begin{cases}
    \displaystyle
    \sum_{k=1}^{K}
    \pi_{i,k}
    \mathcal{N}
    \left(
    x_{i+1};
    \mu_{i,k},
    \sigma_{i,k}^{2}
    \right),
    &
    r_i=1,
    \\[12pt]
    q_{i,x_{i+1}},
    &
    r_i\in\{0,2\}.
    \end{cases}
\]

\subsection{Model Training}
\label{sec:training}

The autoregressive model was trained end-to-end with next-token prediction as its self-supervised objective, jointly optimizing discrete event prediction and continuous value density estimation. 

Let $
\mathcal D
=
\left\{
i:
x_{i+1}\in\mathcal V
\right\}
$
denote the trajectory positions for which the target is discrete, and let
$
\mathcal C
=
\left\{
i:
r_i=1
\right\}
$
denote the positions for which the target is continuous.
For discrete targets, we computed the focal loss $\mathcal L_{\mathrm{FL}}$ defined as

\[
\mathcal L_{\mathrm{FL}}
=
-
\frac{1}{|\mathcal D|}
\sum_{i\in\mathcal D}
\left(
1-q_{i,x_{i+1}}
\right)^\gamma
\log q_{i,x_{i+1}}.
\]

where $\gamma$ is the modulating factor used to reduce the dominance of frequent and easily predicted clinical events.

For continuous targets, we computed the negative log-likelihood $\mathcal L_{\mathrm{NLL}}$, defined as

\[
\mathcal L_{\mathrm{NLL}}
=
-
\frac{1}{|\mathcal C|}
\sum_{i\in\mathcal C}
\log
\left[
\sum_{k=1}^{K}
\pi_{i,k}
\frac{1}{\sqrt{2\pi}\sigma_{i,k}}
\exp
\left(
-\frac{
\left(
x_{i+1}-\mu_{i,k}
\right)^2
}{
2\sigma_{i,k}^{2}
}
\right)
\right].
\]

Overall, the total loss at each training step is

\[
\mathcal{L} = \mathcal{L}_{\text{FL}} + \lambda \, \mathcal{L}_{\text{NLL}}
\]

where $\mathcal{L}_{\text{FL}}$ is the focal loss applied to the language modeling head logits at discrete targets, and $\mathcal{L}_{\text{NLL}}$ is the negative log-likelihood of the ground-truth continuous value under the predicted GMM density weighted by $\lambda$. 

\subsection{Inference and Trajectory Generation}
\label{sec:inference}
 
At inference time, PedsFM generates trajectories autoregressively. For discrete targets, the next token is sampled using multinomial sampling from the temperature-scaled categorical distribution produced by the language modeling head. For continuous targets, the next trajectory element is sampled from the predicted Gaussian mixture distribution. The sampled value is mapped to the original clinical scale, reverting the normalization. Additionally, we implemented a repetition penalty mechanism to prevent the model from becoming trapped in repetitive cycles during long rollouts (see Supplementary Methods).

\paragraph*{Inference Stopping Criteria.} The autoregressive generation ends when one of the stopping criteria is met: 1) maximum number of tokens generated, 2) predefined stopping token encountered, 3) end of sequence token encountered, and 4) time horizon exceeded. For the last criterion, the time progression is calculated using the lower ($t_1$) and upper ($t_2$) bounds of the delta tokens as follows, 

\[
\Delta t = \sqrt{t_1 \times t_2}.
\]

This corresponds to the geometric midpoint of the interval represented by the temporal tokens \cite{waxler_generative_2025}.

\subsection{Model Evaluation}
\label{sec:evaluation}
 
We evaluate the pretrained PedsFM along three complementary axes on the test split: generative plausibility, zero-shot probabilistic prediction of clinically relevant outcomes, and representation learning via linear probing on frozen embeddings. The temperature was set to 0.5 for all evaluation tasks.

\subsubsection{Zero-shot Prediction}

\paragraph*{Monte Carlo Rollout-Based Prediction.} From a fixed anchor point in a patient's trajectory, PedsFM autoregressively generates $N$ continuations by sampling from its predicted token and GMM distributions at each step. The evaluated outcome probability is estimated as the fraction of $N$ simulated rollouts in which the event of interest, represented by a discrete token, occurs. This approach is used for outcomes that unfold over a prediction window and depend on the joint trajectory of many variables, such as early mortality and vasopressor initiation.

\paragraph*{Query-Conditioned GMM Scoring.} Instead of generating multiple future tokens, the patient trajectory is extended with task-specific query tokens, and a single forward pass through PedsFM yields the GMM head parameters for the next trajectory element at that query position. Because the GMM head models a full continuous distribution rather than a point estimate, this single pass is sufficient to compute the expected value, or an upper or lower tail probability for any clinically defined threshold $\tau$, without sampling. It is important to note that this approach is exclusively used for continuous outcomes tied to a single measurement, such as a specific collection of laboratory samples.

Given the predicted Gaussian mixture distribution, the expected standardized value is

\[
\mathbb E[X\mid\mathbf{h}]
=
\sum_{k=1}^{K}
\pi_k\mu_k.
\]

For an upper tail probability prediction, using a defined clinical threshold $\tau$ normalized like the numerical values, an upper-tail probability is

\[
P
\left(
X\geq\tau\mid\mathbf{h}
\right)
=
\sum_{k=1}^{K}
\pi_k
\left[
1-
\Phi
\left(
\frac{
\tau-\mu_k
}{
\sigma_k
}
\right)
\right],
\]

while a lower-tail probability is 

\[
P
\left(
X<\tau^\ast\mid\mathbf{h}
\right)
=
\sum_{k=1}^{K}
\pi_k
\Phi
\left(
\frac{
\tau^\ast-\mu_k
}{
\sigma_k
}
\right),
\]

where $\Phi$ denotes the standard normal cumulative distribution function.

\subsubsection{Representation Learning}

In addition to the generative capabilities evaluated with the zero-shot approach, we assessed the representations learned with PedsFM and whether they were clinically informative. We evaluated two alternative representations: 1) the average of the last hidden state, applying mean-pooling across all sequence positions, and 2) the last hidden state at the final sequence position. The second is the commonly chosen approach in generative EHR foundation models \cite{wornow_ehrshot_2023, guo_tokenization_2026}. 

For a trajectory of length $L$, represented by $L$ hidden states, these representations are

\[
\mathbf{z}
=
\frac{1}{L}
\sum_{i=1}^{L}
\mathbf{h}_i
\]

and

\[
\mathbf{z}
=
\mathbf{h}_L,
\]

where
$\mathbf{z}\in\mathbb R^d$.

These representations were extracted from the frozen pretrained PedsFM, and served as the input to a logistic regression classifier that was trained for each downstream task. Besides a fully supervised linear probe, we assessed sample efficiency by training the logistic regression using progressively larger subsets of the training split.

\subsubsection{Vital Signs Plausibility Analysis}

The plausibility analysis of the vital signs was conducted by comparing the generated vital signs to the actual observed measurements, computing the Pearson correlation $r$ and the mean absolute error (MAE). This assessment used the Monte Carlo rollout-based prediction with $N=10$, using the first 12 hours of hospital admission to generate the following 12 hours and calculate the average for each vital sign. As a reference baseline, we used the last-seen-value carry forward, method commonly used for handling missing data in EHR modeling \cite{harutyunyan_multitask_2019}.

Additionally, we conducted an analysis to assess the learned associations of well characterized and represented medication-vital pairs for a subset of 1,000 hospitalizations from the test split that did not have any medication administrations within the first 12 hours of admission. Specifically, we studied epinephrine-heart rate, and fentanyl-respiratory rate. For each pair, we used the query-conditioned GMM scoring anchored to the first 12 hours of admission to generate the corresponding unconditioned and conditioned vital sign measurement. More specifically, for the unconditioned generation, we appended two query tokens, \texttt{DELTA\_5min-15min},  \texttt{VITAL\_<\textit{vital}>} to the trajectory seed, and calculated the expected value of the evaluated vital, $\mathbb{E}[vital]$, from the generated distribution. Similarly, for the conditioned generation, we followed the same process, but we artificially appended the selected medication tokens, either one or multiple times, before the two query tokens.

It is important to note that this analysis is intended as an internal consistency check of the model, evaluating whether PedsFM learned pharmacologically expected directional shifts, e.g., increased heart rate following epinephrine administration and decreased respiratory rate following fentanyl administration. It does not constitute a study or claim of causal drug effect.

\subsubsection{Downstream Tasks}

We selected tasks representative of three complementary task types: outcome tasks, which predict a terminal or adverse clinical endpoint; intervention tasks, which predict the initiation of a clinical intervention; and physiological tasks, which predict the value of a physiological measurement. More specifically, we evaluated early in-hospital mortality, early vasopressor initiation, and laboratory abnormalities: lactate $\geq 5$ mmol/L, platelets $< 100 \times 10^3/\mu L$, glucose $> 150$ mg/dL, and total bilirubin $\geq 4$ mg/dL. All downstream tasks required a hospital admission of at least 24 hours and only considered events that occurred within the first seven days of admission.

\paragraph*{Early In-Hospital Mortality.} Early in-hospital mortality was defined as death occurring within the first seven days of admission. Such encounters constituted the cases; and the rest, mortality after day seven, short and long stays, were the controls. We used the Monte Carlo rollout-based prediction, estimating mortality probability as the fraction of the generated continuations in which the death token appears, i.e., $N_{expired}/N_{total}$ with $N_{total}=20$. The prediction time was 24 hours after admission, and trajectories were generated until they exceeded a time limit of seven days or a death token was generated. To avoid noisy labels, encounters with a discharge disposition of \textit{end-of-life care} or \textit{unknown} were excluded from this task.

\paragraph*{Early Vasopressor Initiation.} Early vasopressor initiation was defined as an initial administration of epinephrine, norepinephrine, dopamine, phenylephrine, or vasopressin during the first 24 hours of admission. We used the Monte Carlo rollout-based prediction, estimating the probability of early vasopressor initiation as the fraction of $N_{total}=20$ generated continuations in which the vasopressors hierarchical tokens appear. The prediction time was six hours after hospital admission and trajectories were generated until they exceeded a time limit of 18 hours. Additionally, hospitalizations with a vasopressor initiation within the first six hours of admission were excluded from this task.

\paragraph*{Lab Abnormalities.} Lab abnormalities were selected using clinically relevant organ dysfunction thresholds from the Phoenix criteria and expanded Phoenix-8 criteria \cite{sanchez-pinto_development_2024}. Lactate $\geq 5$ mmol/L, platelets $< 100 \times 10^3/\mu L$, glucose $> 150$ mg/dL, and total bilirubin $\geq 4$ mg/dL were evaluated in this task. The glucose threshold corresponds specifically to the upper abnormality threshold of the endocrine component, for which glucose $<50$ mg/dL or $>150$ mg/dL is considered abnormal. We used query-conditioned GMM scoring, computing the upper or lower tail probability directly from the GMM head parameters at the specimen collection timestamp. The query tokens were the delta token \texttt{DELTA\_15min-1h} and a laboratory-specific result token \texttt{LAB\_RESULT\_<\textit{lab}>} appended. All laboratory tests within the first seven days of admission were treated as separate prediction instances anchored to their specimen collection timestamp, which preceded the recorded result by a median of 29 minutes. 

\subsubsection{Baseline Model}

A fully supervised task-specific XGBoost was selected as the baseline model for performance comparison with PedsFM zero-shot performance and learned representations. In previous studies, XGBoost has been demonstrated to be the strongest baseline model for EHR-based tasks \cite{gao2025countbased, renc_zero_2024, waxler_generative_2025}. Since the selected baseline is not suitable for sequential data, we extracted summary features from the clinical trajectories available before the task-specific prediction time. We aggregated the discrete events using counts, following a bag-of-words (BoW) approach, and the numerical values using the mean, median, minimum, maximum, and standard deviation.

\subsection{Statistical Analysis}
\label{sec:statistical}

We evaluated the performance of all the studied models using the Area Under the Receiver Operating Characteristic Curve (AUROC) and the Area Under the Precision-Recall Curve (AUPRC). For the plausibility analysis, we used the Pearson correlation ($r$) and the mean absolute error (MAE). Additionally, for all reported results, 95\% confidence intervals were computed via bootstrap resampling. Finally, non-parametric paired comparisons were conducted using the Wilcoxon signed-rank test, with $p < 0.05$ considered statistically significant.

\begin{table}[h]
\centering
\footnotesize
\caption{Demographic characteristics of the study cohort across the training, validation, and test splits at the hospital encounter level.}
\label{tab:cohort}
\begin{tabular}{llllll}
\toprule
\textbf{Characteristic} & \textbf{Category} &
\textbf{Overall} &
\textbf{Train} &
\textbf{Validation} &
\textbf{Test} \\
\midrule

\textbf{Age, years}
& Median (IQR)
& 4 (1--11)
& 4 (1--11)
& 4 (1--11)
& 4 (1--11) \\

\addlinespace
\multirow[t]{6}{*}{\textbf{Age group}}
& Neonate
& 2,773 (3.6\%)
& 2,366 (3.6\%)
& 139 (3.7\%)
& 268 (3.5\%) \\

& Infant
& 15,159 (19.8\%)
& 12,870 (19.7\%)
& 753 (20.1\%)
& 1,536 (20.0\%) \\

& Toddler
& 14,809 (19.3\%)
& 12,575 (19.3\%)
& 753 (20.1\%)
& 1,481 (19.3\%) \\

& Preschooler
& 10,625 (13.9\%)
& 9,085 (13.9\%)
& 533 (14.2\%)
& 1,007 (13.1\%) \\

& School age
& 15,127 (19.7\%)
& 12,871 (19.7\%)
& 710 (19.0\%)
& 1,546 (20.1\%) \\

& Adolescent
& 18,159 (23.7\%)
& 15,461 (23.7\%)
& 854 (22.8\%)
& 1,844 (24.0\%) \\

\addlinespace
\multirow[t]{3}{*}{\textbf{Sex}}
& Female
& 34,397 (44.9\%)
& 29,216 (44.8\%)
& 1,714 (45.8\%)
& 3,467 (45.1\%) \\

& Male
& 42,252 (55.1\%)
& 36,010 (55.2\%)
& 2,028 (54.2\%)
& 4,214 (54.9\%) \\

& Unknown
& 3 ($<$0.1\%)
& 2 ($<$0.1\%)
& 0 (0.0\%)
& 1 ($<$0.1\%) \\

\addlinespace
\multirow[t]{7}{*}{\textbf{Race}}
& American Indian
& 336 (0.4\%)
& 283 (0.4\%)
& 14 (0.4\%)
& 39 (0.5\%) \\

& Asian
& 2,259 (2.9\%)
& 1,929 (3.0\%)
& 102 (2.7\%)
& 228 (3.0\%) \\

& Black
& 29,391 (38.3\%)
& 24,884 (38.1\%)
& 1,518 (40.6\%)
& 2,989 (38.9\%) \\

& Pacific Islander
& 104 (0.1\%)
& 80 (0.1\%)
& 7 (0.2\%)
& 17 (0.2\%) \\

& White
& 30,822 (40.2\%)
& 26,239 (40.2\%)
& 1,489 (39.8\%)
& 3,094 (40.3\%) \\

& Other
& 6,223 (8.1\%)
& 5,331 (8.2\%)
& 300 (8.0\%)
& 592 (7.7\%) \\

& Unknown
& 7,517 (9.8\%)
& 6,482 (9.9\%)
& 312 (8.3\%)
& 723 (9.4\%) \\

\addlinespace
\multirow[t]{3}{*}{\textbf{Ethnicity}}
& Hispanic
& 10,398 (13.6\%)
& 8,927 (13.7\%)
& 452 (12.1\%)
& 1,019 (13.3\%) \\

& Non-Hispanic
& 62,116 (81.0\%)
& 52,770 (80.9\%)
& 3,092 (82.6\%)
& 6,254 (81.4\%) \\

& Unknown
& 4,138 (5.4\%)
& 3,531 (5.4\%)
& 198 (5.3\%)
& 409 (5.3\%) \\

\midrule
\textbf{Total}
&
& \textbf{76,652}
& \textbf{65,228}
& \textbf{3,742}
& \textbf{7,682} \\

\bottomrule
\end{tabular}
\end{table}

\section{Results}
\label{sec:results}

\subsection{Cohort Characteristics}

The multicenter cohort included 76,652 hospitalizations from 56,655 unique pediatric patients across three hospitals. For patients with multiple hospitalizations, each of them was included as an independent observation. The demographic characteristics are summarized in Table \ref{tab:cohort}, grouped by train, test, and validation splits. The median age at admission was 4 years (IQR, 1–11), and 55.1\% of hospitalizations were male patients. Additionally, all pediatric age groups were represented, with adolescents contributing the largest group (23.7\%) of hospitalizations and neonates the smallest (3.6\%).

Patient-level partitioning yielded 65,228 training, 3,742 validation, and 7,682 test hospitalizations. Additionally, harmonizing and tokenizing the EHR data to obtain the clinical trajectories resulted in a total of 280.65 million tokens, of which 237.41 million were used to train the model, and a vocabulary of 510 unique discrete tokens.

\subsection{PedsFM Training and Inference}
\label{sec:training}

The selected PedsFM configuration consisted of 12 transformer decoder layers, 12 attention heads, a hidden dimension of 768, and a context size of 8,192, for a total of 86.57 million parameters (see architecture details and training specifications in Supplementary Table \ref{tab:specifications}). Additionally, FlashAttention was used to improve efficiency. The model was trained for approximately 48 hours on two NVIDIA H200 GPUs using distributed data parallelism (DDP). At inference time, autoregressive generation required approximately 4 seconds per 1,000 generated tokens on a single NVIDIA H200 GPU.

To assess whether the model learned clinical structure in its discrete token vocabulary, we visualized the learned embedding vectors for all discrete tokens using UMAP (see Supplemental Figure \ref{fig:embeddings}). Despite being trained only on self-supervised next-token prediction,  the model learned token representations that clustered by clinical category, with medication tokens forming the largest and most distinct cluster, and laboratory-related tokens (lab results, specimen collection) forming a separate, adjacent cluster alongside vital signs. Additionally, for some categories like demographics and imaging procedures, distinct clusters formed within the category based on semantic similarity, e.g., clusters for the different types of demographic data and age groups. This separation suggests that the model organizes its discrete vocabulary along clinically meaningful axes without any explicit category supervision.

\subsection{Vital Signs Plausibility and Pharmacological Associations}
\label{sec:vitals}

We assessed the physiological plausibility of the generated trajectories by generating the next 12 hours of the clinical trajectory, anchored at hour 12 after hospital admission. In general, the generated vital signs are consistent with the observed measurements. Compared to the last-seen-value carry-forward baseline, PedsFM generations showed comparable or higher correlation and lower MAE for heart rate ($r=$ 0.88 vs. 0.85; MAE $=$ 9.73 vs. 11.90 bpm), respiratory rate ($r=$ 0.83 vs.\ 0.81; MAE $=$ 4.34 vs. 4.96 breaths/min), systolic blood pressure, diastolic blood pressure, mean arterial pressure (MAP), and temperature (see Supplementary Table \ref{tab:vital_analysis}). SpO$_2$ was the only vital sign for which the carry-forward baseline showed better agreement with observed values, likely reflecting the narrow, ceiling-skewed distribution of this variable, which favors a naive approach such as carry-forward. In general, these results support the physiological plausibility of the generated measurements.

\begin{figure}[h]
    \centering
    \includegraphics[width=1.0\linewidth]{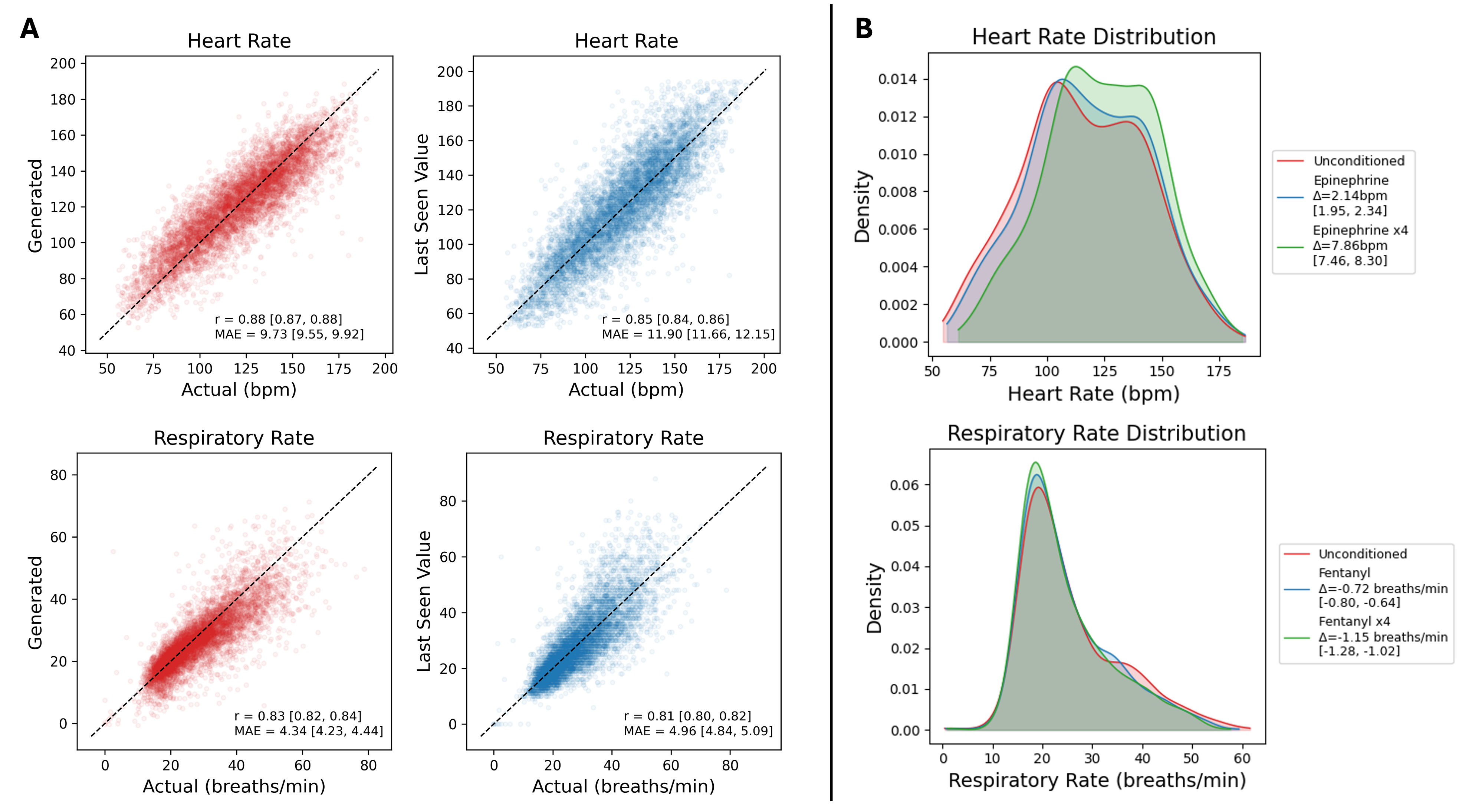}
    \caption{Vital sign generation analysis and learned associations assessment results.
A) Scatter plots of generated vital signs vs. actual observed measurements for heart rate (top) and respiratory rate (bottom), comparing Monte Carlo rollout (left, red) to the last-seen-value carry-forward baseline (right, blue). Dashed line indicates perfect agreement $(y = x)$. Pearson correlation (r) and mean absolute error (MAE) are reported. B) Generated heart rate (top) and respiratory rate (bottom) distributions under unconditioned generation (red) vs. generation conditioned on a medication (blue) or four-fold conditioning (green) of the corresponding medication. $\Delta$ indicates the shift in mean vital sign value relative to the unconditioned distribution.}
    \label{fig:vitals}
\end{figure}

The medication-conditioning analysis also produced directionally expected shifts in the predicted vitals distributions (see Figure \ref{fig:vitals}). Conditioning generation on epinephrine administration increased the predicted heart rate by an average of 2.14 bpm (95\% CI 1.95, 2.34 bpm). Repeated epinephrine conditioning, assessed by adding four consecutive administration token sets, increased it by 7.86 bpm (7.46, 8.30 bpm). For the respiratory rate, conditioning on fentanyl shifted the predictions downward, with an average decrease of 0.72 breaths/min ($-$0.80, $-$0.64 breaths/min) and 1.15 breaths/min ($-$1.28, $-$1.02 breaths/min) under repeated conditioning. Because the medication dose was not represented in the trajectories, these results should be interpreted as sensitivity to repeated medication conditioning rather than as a pharmacologic dose response relationship.  

Interestingly, the epinephrine-associated shift was more pronounced at lower baseline heart rate values, whereas the fentanyl-associated shift was greater at higher baseline respiratory rates. These baseline-dependent responses are physiologically plausible, as patients with lower heart rates have greater capacity for a detectable response, while fentanyl-induced respiratory suppression may be more evident among patients with initially elevated respiratory rates. This evaluation was statistically significant with $p < 0.01$ for both vitals.

\subsection{Downstream Tasks Performance}
\label{sec:downstream}

Zero-shot results demonstrated performance above chance across all evaluated tasks for both AUROC and AUPRC. For early in-hospital mortality and vasopressor initiation, zero-shot Monte Carlo rollout achieved an AUROC of 0.868 and an AUPRC of 0.260, and an AUROC of 0.814 and an AUPRC of 0.643, respectively. The fully supervised XGBoost models achieved higher performance for both tasks, with AUROC and AUPRC values of 0.952 and 0.627 for mortality, and 0.949 and 0.807 for vasopressor initiation (see Table \ref{tab:results}). Nevertheless, the zero-shot results showed that PedsFM could identify patients at increased risk without task-specific training (see Supplementary Figure \ref{fig:stratified_mort}). For the four laboratory abnormality tasks, evaluated using query-conditioned GMM scoring, zero-shot outperformed a fully supervised XGBoost with AUROC/AUPRC values of 0.954/0.912 for lactate, 0.976/0.907 for platelets, 0.988/0.935 for total bilirubin, and 0.877/0.664 for glucose. The strongest discrimination was observed for bilirubin and platelets, while glucose remained the most challenging laboratory outcome. The differences between the rollout-based outcome tasks and the query-conditioned laboratory tasks reflect some methodological differences. Extended autoregressive rollouts introduce sampling variability, and the prediction errors propagate and accumulate through the generated trajectories. On the other hand, query-conditioned GMM scoring obtain the required conditional distribution in a single forward pass. However, it is not possible to attribute the performance differences exclusively to the inference strategy because the evaluated tasks also differ in prevalence, prediction horizon, target type, and alignment with the pretraining objective.

\begin{table}[h]
\centering
\footnotesize
\caption{Comparison of zero-shot PedsFM performance and a fully supervised task-specific XGBoost baseline on downstream prediction tasks.}
\label{tab:results}

\begin{tabular}{lcccc}
\toprule
& \multicolumn{2}{c}{\textbf{Zero-shot PedsFM}}
& \multicolumn{2}{c}{\textbf{XGBoost}} \\
\cmidrule(lr){2-3}
\cmidrule(lr){4-5}

\textbf{Prediction Task}
& \textbf{AUROC}
& \textbf{AUPRC}
& \textbf{AUROC}
& \textbf{AUPRC} \\
\midrule

\begin{tabular}[c]{@{}l@{}}
Early In-hospital Mortality \\
\footnotesize Prevalence: 1.0\%
\end{tabular}
&
\begin{tabular}[c]{@{}c@{}}
0.868 \\
{}[0.812--0.919]
\end{tabular}
&
\begin{tabular}[c]{@{}c@{}}
0.260 \\
{}[0.167--0.370]
\end{tabular}
&
\begin{tabular}[c]{@{}c@{}}
0.960 \\
{}[0.937--0.980]
\end{tabular}
&
\begin{tabular}[c]{@{}c@{}}
0.594 \\
{}[0.470--0.709]
\end{tabular}
\\
\addlinespace

\begin{tabular}[c]{@{}l@{}}
Early Vasopressor Initiation \\
\footnotesize Prevalence: 5.7\%
\end{tabular}
&
\begin{tabular}[c]{@{}c@{}}
0.814 \\
{}[0.786--0.841]
\end{tabular}
&
\begin{tabular}[c]{@{}c@{}}
0.643 \\
{}[0.598--0.687]
\end{tabular}
&
\begin{tabular}[c]{@{}c@{}}
0.947 \\
{}[0.933--0.957]
\end{tabular}
&
\begin{tabular}[c]{@{}c@{}}
0.790 \\
{}[0.754--0.823]
\end{tabular}
\\
\addlinespace

\begin{tabular}[c]{@{}l@{}}
Abnormal Lactate \\
\footnotesize Prevalence: 30.3\%
\end{tabular}
&
\begin{tabular}[c]{@{}c@{}}
0.954 \\
{}[0.945--0.962]
\end{tabular}
&
\begin{tabular}[c]{@{}c@{}}
0.912 \\
{}[0.895--0.926]
\end{tabular}
&
\begin{tabular}[c]{@{}c@{}}
0.930 \\
{}[0.916--0.940]
\end{tabular}
&
\begin{tabular}[c]{@{}c@{}}
0.874 \\
{}[0.849--0.896]
\end{tabular}
\\
\addlinespace

\begin{tabular}[c]{@{}l@{}}
Abnormal Platelets \\
\footnotesize Prevalence: 17.1\%
\end{tabular}
&
\begin{tabular}[c]{@{}c@{}}
0.976 \\
{}[0.973--0.979]
\end{tabular}
&
\begin{tabular}[c]{@{}c@{}}
0.907 \\
{}[0.896--0.917]
\end{tabular}
&
\begin{tabular}[c]{@{}c@{}}
0.959 \\
{}[0.956--0.963]
\end{tabular}
&
\begin{tabular}[c]{@{}c@{}}
0.852 \\
{}[0.840--0.865]
\end{tabular}
\\
\addlinespace

\begin{tabular}[c]{@{}l@{}}
Abnormal Bilirubin \\
\footnotesize Prevalence: 10.0\%
\end{tabular}
&
\begin{tabular}[c]{@{}c@{}}
0.988 \\
{}[0.984--0.991]
\end{tabular}
&
\begin{tabular}[c]{@{}c@{}}
0.935 \\
{}[0.922--0.947]
\end{tabular}
&
\begin{tabular}[c]{@{}c@{}}
0.979 \\
{}[0.974--0.984]
\end{tabular}
&
\begin{tabular}[c]{@{}c@{}}
0.902 \\
{}[0.886--0.916]
\end{tabular}
\\
\addlinespace

\begin{tabular}[c]{@{}l@{}}
Abnormal Glucose \\
\footnotesize Prevalence: 18.4\%
\end{tabular}
&
\begin{tabular}[c]{@{}c@{}}
0.877 \\
{}[0.871--0.883]
\end{tabular}
&
\begin{tabular}[c]{@{}c@{}}
0.664 \\
{}[0.647--0.680]
\end{tabular}
&
\begin{tabular}[c]{@{}c@{}}
0.835 \\
{}[0.828--0.842]
\end{tabular}
&
\begin{tabular}[c]{@{}c@{}}
0.591 \\
{}[0.574--0.608]
\end{tabular}
\\
\addlinespace

\bottomrule
\end{tabular}
\end{table}

Additionally, we evaluated whether representations learned by PedsFM were clinically informative when used as input to a linear probe (see Supplementary Figure \ref{fig:rep_results}). For some of the evaluated tasks, linear probes trained on frozen PedsFM embeddings approached the performance of the corresponding fully supervised XGBoost baseline. Interestingly, mean-pooled representations consistently outperformed the last-token representation across all tasks, which suggests that information relevant to downstream prediction was distributed throughout the dense clinical sequence rather than being fully summarized by its final token despite the causal attention architecture. For early in-hospital mortality and vasopressors initiation prediction, linear probes on PedsFM representations achieved performance comparable to the fully supervised XGBoost baseline and substantially higher than zero-shot rollout prediction, particularly in AUPRC. Additionally, linear probes outperformed XGBoost on limited training samples, suggesting sample efficiency gains introduced by the pretrained model. In contrast, the linear probes improved more gradually for the laboratory tasks, and remained below both query-conditioned GMM scoring and XGBoost, even with all labeled training samples. These results suggest that aggregated representations effectively summarize the broader patient state associated with clinical outcomes and interventions, whereas laboratory abnormalities are more directly captured by the generative continuous-value pathway. Therefore, the GMM head was better suited to representing the measurement-specific and threshold-dependent variation that was not easily separable by the logistic regression classifier applied to the latent trajectory representation.

\section{Discussion}
\label{sec:discussion}

This study introduces PedsFM, a pediatric-specific generative foundation model that jointly generates discrete clinical events and continuous physiological measurements. Using a single transformer decoder architecture that leverages deterministic pathways, the model learned from critical care clinical trajectories, which are dense and heterogeneous, while preserving continuous physiological measurements at their original numerical resolution. Therefore, there are two main differences between PedsFM and most existing EHR foundation models. First, the numerical values in our trajectories do not require discretization, which is particularly important in pediatrics, where the interpretation of a numerical value can vary substantially with age \cite{lyle_current_2022}. Second, unlike existing models that heavily rely on diagnoses, procedures, medication orders, and other relatively sparse events, we preserved dense physiological information by including the vital signs \cite{guo_systematic_2026, wornow_ehrshot_2023}. We trained and evaluated the model on data from three hospitals with pediatric-specific facilities, and we centered our evaluation around the prediction of relevant clinical events that could potentially provide actionable insights to clinicians. Despite focusing on the generative capabilities of the model, we also assessed the representations learned during the pretraining for multiple downstream tasks. Additionally, we evaluated the model using complementary approaches, such as plausibility of vital signs and medication–physiology associations, that allowed us to have a broader assessment than discrimination metrics alone.

The physiological evaluation showed that PedsFM learned clinically meaningful conditional relationships rather than simply memorizing common vital sign values present in the data. Generated heart rate, respiratory rate, blood pressure, and temperature measurements were overall more consistent with subsequently observed values than a last-seen-value carry-forward baseline. The exception was SpO$_2$, for which the narrow and ceiling-skewed distribution may have favored the carry-forward baseline. The inclusion of additional, complementary distributions in the mixture density will be studied in future work as it may be useful to model some of the skewed or physiologically bounded continuous variables. Conditioning the model on epinephrine shifted the predicted heart rate distribution upward, whereas conditioning it on fentanyl shifted the predicted respiratory rate distribution slightly downward \cite{motiejunaite_adrenergic_2021, saunders_kolliker-fuseparabrachial_2020}. The larger shifts observed under repeated medication conditioning, together with their dependence on the baseline physiological state, suggest that the model captured contextual and nonlinear associations between medications and physiology. Hence, these experiments constitute an important model consistency and plausibility check that could be adopted as a complementary evaluation to standard discriminatory metrics. However, it is important to note that this is not a counterfactual analysis and does not imply a causal effect since the model was trained on observational data and multiple confounding factors may exist.

Additionally, in evaluating our model, we identified a contrast between how zero-shot generations and linear-probing performed across downstream tasks. Linear probes on frozen PedsFM representations achieved comparable performance to XGBoost for early mortality and vasopressors initiation prediction, but remained below zero-shot query-conditioned GMM scoring for all four laboratory abnormality tasks, even with the full training set available. We interpret this not as a weakness of the representation but as an indication that clinical tasks of different nature may demand different decoding mechanisms and clinical context for dense and heterogeneous trajectories like the ones observed in the PICU. In this case, mortality and vasopressors initiation are macro, cumulative endpoints that reflect an accumulation of information across physiological systems, interventions, and time, besides having a longer prediction horizon. These outcomes may be effectively represented by vector embeddings that summarize the patient state, and are separable by a linear boundary. On the other hand, laboratory values are dynamic, continuous variables, and the associated prediction task, i.e., prediction of abnormal laboratory results, is threshold-dependent, has a shorter and more uniform prediction horizon, and it is more aligned with the pretraining objective. Query-conditioned GMM scoring preserves the measurement-specific context and the full conditional distribution at the relevant position in the sequence, resulting in a strong performance for the four laboratory tasks. Similarly, the measurement-specific context is explicitly included in the features extracted to train the XGBoost baseline, which results in increased performance in these tasks when compared to the representations. Nonetheless, the zero-shot rollout strategy achieved a substantial fold improvement over the prevalence baseline without additional training for early in-hospital mortality and vasopressors initiation. The model assigned progressively higher predicted risk across clinically meaningful outcome groups, with the highest risk observed among patients who died within the first seven days of admission, followed by those who died after seven days, patients with prolonged hospital stays, and finally patients with short hospital stays.

An interesting finding derived from our downstream task evaluations was that mean-pooled representations consistently outperformed last-token representations. This is an important difference from previous work. Existing decoder-only clinical foundation models generally extract the final hidden state as the patient-level representation \cite{wornow_ehrshot_2023, fallahpour_ehrmamba_2024}. Because causal attention allows the final position to attend to all preceding tokens, it is assumed that it effectively summarizes the clinical trajectory. However, our results suggest that, for dense longitudinal EHR trajectories, aggregating information across the entire sequence preserves relevant information for the downstream tasks that was not captured as effectively by the last token representation. Alternative strategies such as learned attention-pooling, temporally weighted aggregation, or hierarchical trajectory representations may further improve the extraction of the patient state in future work.

It is important to interpret these results in light of the dataset scale. Recent large-scale EHR foundation models, such as Curiosity \cite{waxler_generative_2025}, trained on Epic Cosmos data that contains over 150 billion tokens across multiple healthcare institutions, demonstrated that increasing model and training dataset size yielded better results for different downstream tasks, consistent with scaling behavior first characterized in general-purpose language models \cite{kaplan_scaling_2020}. The dataset created in this work is several orders of magnitude below Cosmos, with 237 million training tokens drawn from approximately 65,000 hospitalizations. PICU admissions in particular are rarer and more heterogeneous across developmental stages than adult hospitalizations, concentrated in a smaller number of specialized centers, and overall, children are severely underrepresented in publicly available datasets \cite{hua_underrepresentation_2026}. Given the previously studied scaling relationships, it is possible that our zero-shot rollout performance on long horizon, low prevalence outcomes, such as in-hospital mortality reflects this data ceiling. However, with our pediatric-specific dataset, we demonstrated the potential and strengths of the proposed architecture for different critical care applications. Currently, we are working on the creation of an extensive pediatric corpora that helps close this data gap. Nonetheless, pediatric resources will remain constrained in terms of size relative to adult datasets.

This work has some limitations. The cohort, while multisite, remains limited to three hospitals, and independent external validation is required to assess generalizability to institutions with different populations, workflows, and intervention practices. In addition, as mentioned above, our dataset is size-constrained relative to existing EHR foundation model studies. We expect to mitigate these limitations with data from additional pediatric sites, which is an ongoing effort. Also, we exclusively included structured EHR data in our trajectories; however, leveraging the routing mechanism combined with alternative tokenization strategies, additional modalities could be included in the clinical trajectories to have a more complete representation of the patient. Finally, although PedsFM preserves the native resolution of physiologic measurements, it discretizes the temporal component. In future work, we will explore alternative approaches to encode the clinical event times in the trajectories to achieve more granularity and precision.

\section{Conclusion}
\label{sec:conclusion}

PedsFM presents a unified approach to jointly model discrete and continuous clinical events, enabling the extraction of clinically useful information from critical care trajectories without requiring task-specific training. This work could help the advance the progress of EHR modeling in pediatrics, and encourage the development of adaptable and scalable decision support tools. External validation and prospective evaluation are necessary to evaluate the performance of the model in providing clinically actionable insights.

\section*{Data Availability}

The studied dataset is not publicly available due to patient privacy restrictions. 

\section*{Author Contributions}

R.K. and D.C. conceived the study. D.C. designed the methodology, processed the data, developed the foundation model, conducted the experiments, analyzed the results, drafted and edited the manuscript. B.A, M.A., T.I., and K.N. assisted with the evaluation framework, interpretation of the results, and edited the manuscript. H.F and N.S assisted with the clinical interpretation and edited the manuscript. A.R., C.H., I.K., J.M., M.W., A.K., S.C., J.L., C.R., S.B., and R.K. assisted with interpreting the data and edited the manuscript. All authors approved the final version of this manuscript.

%Bibliography
\bibliographystyle{unsrt}  
\bibliography{references}  

\newpage
\section*{Supplementary Material}

\setcounter{table}{0}
\renewcommand{\thetable}{S\arabic{table}}
\renewcommand{\tablename}{Supplementary Table}

\setcounter{figure}{0}
\renewcommand{\thefigure}{S\arabic{figure}}
\renewcommand{\figurename}{Supplementary Figure}

\renewcommand{\thesubsection}{\Alph{subsection}}
\setcounter{subsection}{0}

\subsection{Supplementary Methods}
\label{sec:sup_methods}

\subsubsection{Tokenization Details}

Our tokenization strategy, combined with our data harmonization pipeline, attempted to minimize the number of unique tokens required to represent clinical events and patient trajectories, and to facilitate cross-institution adoption. This strategy resulted in a discrete vocabulary size of 510 unique tokens, and infinite numeric values. See Supplementary Table \ref{tab:tokens} for event-specific tokenization structure.

\begin{table}[H]
\centering
\scriptsize
\caption{PedsFM tokenization.}
\label{tab:tokens}

\begin{tabular}{lcl}
\toprule
\textbf{Clinical Event Type} & \textbf{Number of Tokens} & \textbf{Tokenization} \\
\midrule

\begin{tabular}[t]{@{}l@{}}
\textbf{Demographic}\\
\hspace{1em}Age\\
\hspace{1em}Age Group\\
\hspace{1em}Sex\\
\hspace{1em}Race\\
\hspace{1em}Ethnicity
\end{tabular}
&
\begin{tabular}[t]{@{}l@{}}
\\
1\\
1\\
1\\
1\\
1
\end{tabular}
&
\begin{tabular}[t]{@{}l@{}}
\\
\scriptsize{\texttt{AGE\_YEARS\_<\textit{age\_years}>}} \\
\scriptsize{\texttt{AGE\_GROUP\_<\textit{age\_group}>}} \\
\scriptsize{\texttt{SEX\_<\textit{sex}>}}\\
\scriptsize{\texttt{RACE\_<\textit{race}>}}\\
\scriptsize{\texttt{ETHNICITY\_<\textit{ethnicity}>}}
\end{tabular}
\\
\midrule

\begin{tabular}[t]{@{}l@{}}
\textbf{ED Encounter}\\
\hspace{1em}Admission\\
\hspace{1em}Discharge
\end{tabular}
&
\begin{tabular}[t]{@{}l@{}}
\\
1\\
1
\end{tabular}
&
\begin{tabular}[t]{@{}l@{}}
\\
\scriptsize{\texttt{ED\_ADMISSION}} \\
\scriptsize{\texttt{ED\_DISCHARGE}} \\
\end{tabular}
\\
\midrule

\begin{tabular}[t]{@{}l@{}}
\textbf{Hospital Encounter}\\
\hspace{1em}Admission\\ \\
\hspace{1em}Discharge
\end{tabular}
&
\begin{tabular}[t]{@{}l@{}}
\\
2 \\ \\
2 \\
\end{tabular}
&
\begin{tabular}[t]{@{}l@{}}
\\
\scriptsize{\texttt{HOSPITAL\_ADMISSION}} \\
\scriptsize{\texttt{INSURANCE\_<\textit{insurance}>}} \\
\scriptsize{\texttt{HOSPITAL\_DISCHARGE}} \\ 
\scriptsize{\texttt{DISPOSITION\_<\textit{disposition}>}}
\end{tabular}
\\
\midrule

\begin{tabular}[t]{@{}l@{}}
\textbf{ICU Encounter}\\
\hspace{1em}Admission\\ \\
\hspace{1em}Discharge
\end{tabular}
&
\begin{tabular}[t]{@{}l@{}}
\\
2 \\ \\
1
\end{tabular}
&
\begin{tabular}[t]{@{}l@{}}
\\
\scriptsize{\texttt{ICU\_ADMISSION}} \\
\scriptsize{\texttt{TYPE\_<\textit{icu\_type}>}} \\
\scriptsize{\texttt{ICU\_DISCHARGE}} \\
\end{tabular}
\\
\midrule

\begin{tabular}[t]{@{}l@{}}
\textbf{Unit Transfer}\\
\hspace{1em}Admission\\
\hspace{1em}Discharge
\end{tabular}
&
\begin{tabular}[t]{@{}l@{}}
\\
1\\
1
\end{tabular}
&
\begin{tabular}[t]{@{}l@{}}
\\
\scriptsize{\texttt{<\textit{UNIT}>\_ADMISSION}} \\
\scriptsize{\texttt{<\textit{UNIT}>\_DISCHARGE}} \\
\end{tabular}
\\
\midrule

\begin{tabular}[t]{@{}l@{}}
\textbf{Procedure}\\
\hspace{1em}Respiratory \\ \\
\hspace{1em}Imaging
\end{tabular}
&
\begin{tabular}[t]{@{}l@{}}
\\
2 \\ \\
2 \\
\end{tabular}
&
\begin{tabular}[t]{@{}l@{}}
\\
\scriptsize{\texttt{<\textit{PROCEDURE}>\_placement}} \\
\scriptsize{\texttt{<\textit{PROCEDURE}>\_removal}} \\
\scriptsize{\texttt{IMG\_<\textit{modality}>}} \\ 
\scriptsize{\texttt{SITE\_<\textit{body\_part}>}}
\end{tabular}
\\
\midrule

\begin{tabular}[t]{@{}l@{}}
\vspace{0pt}
\textbf{Medication}
\end{tabular}
& 5-6
&
\begin{tabular}[t]{@{}l@{}}
\scriptsize{\texttt{MED\_<\textit{action}>}} \\
\scriptsize{\texttt{ROUTE\_<\textit{administration\_route}>}} \\
\scriptsize{\texttt{ATC\_MAIN\_<\textit{main\_group}>}} \\
\scriptsize{\texttt{ATC\_THER\_<\textit{therapeutic\_subgroup}>}} \\
\scriptsize{\texttt{ATC\_PHARM\_<\textit{pharmacological\_subgroup}>}} \\
\scriptsize{\texttt{ATC\_chem\_<\textit{chemical\_component}>}} 
\end{tabular}
\\
\midrule

\begin{tabular}[t]{@{}l@{}}
\vspace{0pt}
\textbf{Laboratory Specimen Collection}
\end{tabular}
& 1
& \scriptsize{\texttt{LAB\_TAKEN\_<\textit{lab\_name}>}} 
\\
\midrule

\begin{tabular}[t]{@{}l@{}}
\vspace{0pt}
\textbf{Laboratory Result}
\end{tabular}
& 2
&
\begin{tabular}[t]{@{}l@{}}
\scriptsize{\texttt{LAB\_RESULT\_<\textit{lab\_name}>}} \\
\scriptsize{<\texttt{value}>}
\end{tabular}
\\
\midrule

\begin{tabular}[t]{@{}l@{}}
\vspace{0pt}
\textbf{Vital Sign}
\end{tabular}
& 2
&
\begin{tabular}[t]{@{}l@{}}
\scriptsize{\texttt{VITAL\_<\textit{vital\_name}>}} \\
\scriptsize{<\texttt{value}>}
\end{tabular}
\\
\midrule

\begin{tabular}[t]{@{}l@{}}
\vspace{0pt}
\textbf{Mechanical Ventilation Setting}
\end{tabular}
& 2
&
\begin{tabular}[t]{@{}l@{}}
\scriptsize{\texttt{SETTING\_<\textit{setting\_name}>}} \\
\scriptsize{<\texttt{value}>}
\end{tabular}
\\
\midrule

\begin{tabular}[t]{@{}l@{}}
\vspace{0pt}
\textbf{Body Measurement}
\end{tabular}
& 2
&
\begin{tabular}[t]{@{}l@{}}
\scriptsize{\texttt{BODY\_MEASUREMENT\_<\textit{measurement\_name}>}} \\
\scriptsize{<\texttt{value}>}
\end{tabular}
\\
\midrule

\begin{tabular}[t]{@{}l@{}}
\textbf{Temporal}\\
\hspace{1em}Sequence Start\\
\hspace{1em}Interval
\end{tabular}
&
\begin{tabular}[t]{@{}l@{}}
\\
1\\
1
\end{tabular}
&
\begin{tabular}[t]{@{}l@{}}
\\
\scriptsize{\texttt{TS\_START}} \\
\scriptsize{\texttt{DELTA\_<\textit{interval\_bucket}>}} \\
\end{tabular}
\\

\bottomrule
\end{tabular}
\end{table}

\subsubsection{Inference Repetition Penalty}

During inference, at each generation step, the logits of discrete tokens that recently appeared within a selected window, i.e., the most recent $x$ positions, were down-weighted by a selected factor $>0$, reducing their sampling probability. Temporal tokens were excluded from the penalty since their recurrence is required to represent the intervals between consecutive clinical events. This repetition penalty discouraged the model from repeatedly sampling the same non-temporal discrete tokens, mitigating degenerate loops.

\subsection{Supplementary Results}
\label{sec:sup_results}

\subsubsection{PedsFM Specifications}

Architecture configuration and training specifications are included in Supplementary Table \ref{tab:specifications}.

\begin{table}[H]
\centering
\footnotesize
\caption{PedsFM architecture and training configuration.}
\label{tab:specifications}
\begin{tabular}{ll}
\toprule
\textbf{Specification} & \textbf{Value} \\
\midrule
Transformer layers & 12 \\
Attention heads & 12 \\
Hidden dimension & 768 \\
Feed-forward dimension & 3,072 \\
GMM components & 12 \\
Dropout & 0.20 \\
Maximum context length & 8,192 tokens \\
Total parameters & 86.57 million \\
\midrule
Discrete-token loss & Focal loss ($\gamma=2.0$) \\
Continuous-value loss & NLL \\
Optimizer & AdamW \\
Initial learning rate & $5 \times 10^{-5}$ \\
Batch size & 32 \\
Learning-rate scheduler & Reduce on Plateau \\
Gradient clipping norm & 1.0 \\

\bottomrule
\end{tabular}
\end{table}

\subsubsection{Discrete Token Embedding Space}

UMAP visualization of the discrete embedding vectors for 25 neighbors, minimum distance between points of 0.1, and cosine as the distance metric are depicted in Supplementary Figure \ref{fig:embeddings}. Clusters were consistent across different UMAP configurations.

\begin{figure}[H]
    \centering
    \includegraphics[width=1.0\linewidth]{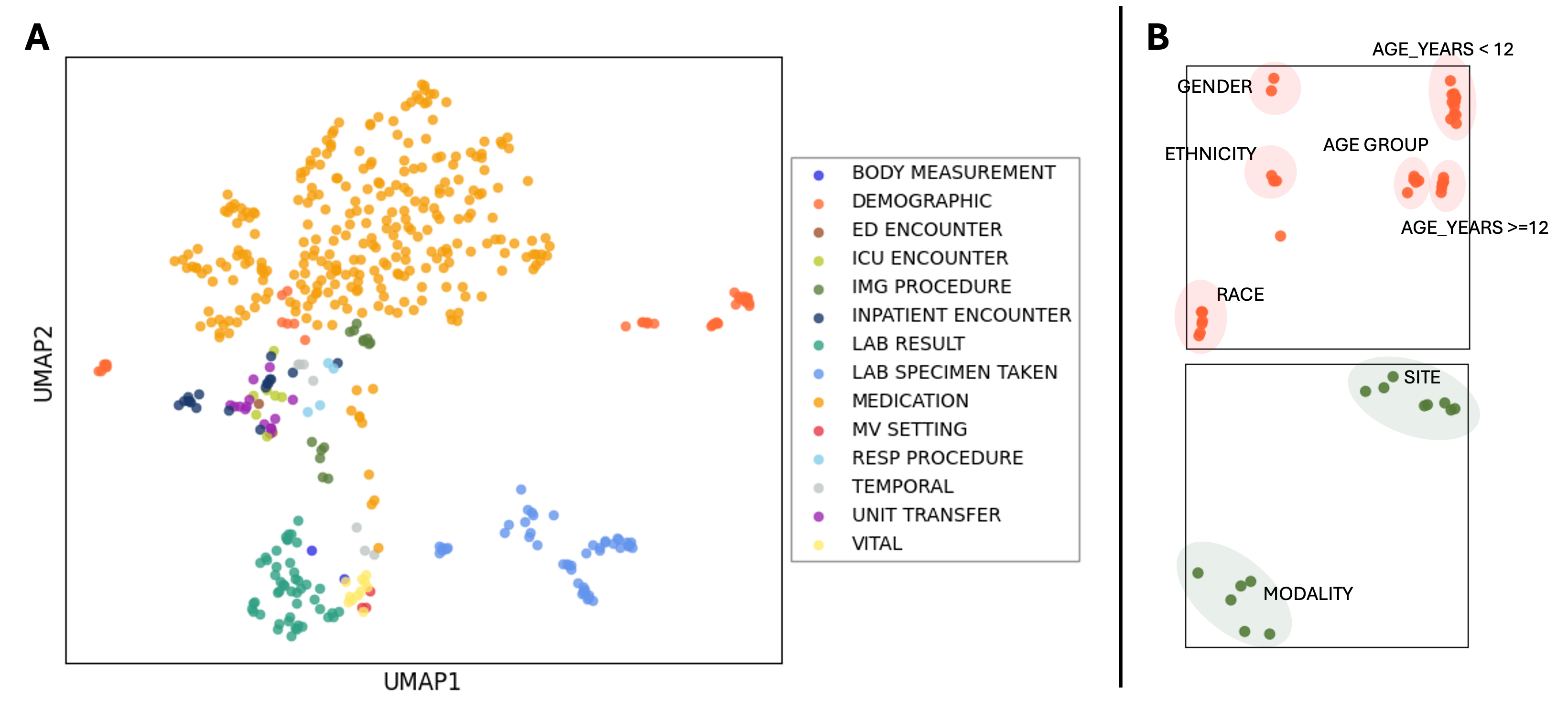}
    \caption{A) UMAP projection of the learned embedding vectors for all discrete vocabulary tokens, colored by token category, e.g., vital sign, lab result, medication, demographic, encounter type). B) Individual visualization of two token subgroups, demographic tokens (top) and imaging procedures (bottom).}
    \label{fig:embeddings}
\end{figure}

\subsubsection{Vital Signs Plausibility Analysis}

Pearson correlation ($r$) and mean absolute error (MAE) for all generated vs. observed vitals are included in Supplementary Table \ref{tab:vital_analysis}.

\begin{table}[H]
\centering
\footnotesize
\caption{Correlation analysis results of generated vital sign predictions and last-seen-value baseline comparison. Values are reported with 95\% CI.}
\label{tab:vital_analysis}
\begin{tabular}{lcccc}
\toprule
& \multicolumn{2}{c}{\textbf{Generated}} &
\multicolumn{2}{c}{\textbf{Last-Seen-Value}} \\
\cmidrule(lr){2-3} \cmidrule(lr){4-5}
\textbf{Vital Sign (Units)} &
$\mathbf{r}$ &
\textbf{MAE} &
$\mathbf{r}$ &
\textbf{MAE} \\
\midrule

Heart Rate (bpm)
&
\begin{tabular}[c]{@{}c@{}}
0.88 \\
{}[0.87--0.88]
\end{tabular}
& 
\begin{tabular}[c]{@{}c@{}}
9.73 \\
{}[9.55--9.92]
\end{tabular}
& 
\begin{tabular}[c]{@{}c@{}}
0.85 \\ 
{}[0.84--0.86]
\end{tabular}
& 
\begin{tabular}[c]{@{}c@{}}
11.90 \\ 
{}[11.66--12.17] 
\end{tabular}
\\
\addlinespace

Respiratory Rate (breaths/min)
& 
\begin{tabular}[c]{@{}c@{}}
0.83 \\ 
{}[0.82--0.84]
\end{tabular}
& 
\begin{tabular}[c]{@{}c@{}}
4.34 \\
{}[4.23--4.44]
\end{tabular}
& 
\begin{tabular}[c]{@{}c@{}}
0.81 \\ 
{}[0.80--0.82]
\end{tabular}
& 
\begin{tabular}[c]{@{}c@{}}
4.96 \\ 
{}[4.84--5.09] 
\end{tabular}
\\
\addlinespace

Systolic Blood Pressure (mmHg)
& 
\begin{tabular}[c]{@{}c@{}}
0.76 \\
{}[0.75--0.77]
\end{tabular}
& 
\begin{tabular}[c]{@{}c@{}}
6.66 \\ 
{}[6.52--6.81]
\end{tabular}
& 
\begin{tabular}[c]{@{}c@{}}
0.70 \\
{}[0.69--0.72]
\end{tabular}
& 
\begin{tabular}[c]{@{}c@{}}
8.74 \\
{}[8.56--8.92]
\end{tabular}
\\
\addlinespace

Diastolic Blood Pressure (mmHg)
& 
\begin{tabular}[c]{@{}c@{}}
0.65 \\
{}[0.63--0.67]
\end{tabular}
& 
\begin{tabular}[c]{@{}c@{}}
6.80 \\
{}[6.67--6.94]
\end{tabular}
& 
\begin{tabular}[c]{@{}c@{}}
0.56 \\
{}[0.54--0.58]
\end{tabular}
& 
\begin{tabular}[c]{@{}c@{}}
9.46 \\
{}[9.25--9.66] 
\end{tabular}
\\
\addlinespace

MAP (mmHg)
& 
\begin{tabular}[c]{@{}c@{}}
0.69 \\ 
{}[0.67--0.70]
\end{tabular}
& 
\begin{tabular}[c]{@{}c@{}}
6.39 \\ 
{}[6.26--6.51]
\end{tabular}
& 
\begin{tabular}[c]{@{}c@{}}
0.60 \\ 
{}[0.59--0.62]
\end{tabular}
& 
\begin{tabular}[c]{@{}c@{}}
8.50 \\ 
{}[8.32--8.68] 
\end{tabular}
\\
\addlinespace

SpO$_2$ (\%)
& 
\begin{tabular}[c]{@{}c@{}}
0.38 \\ 
{}[0.33--0.43]
\end{tabular}
& 
\begin{tabular}[c]{@{}c@{}}
2.01 \\
{}[1.95--2.08]
\end{tabular}
& 
\begin{tabular}[c]{@{}c@{}}
0.60 \\
{}[0.57--0.64]
\end{tabular}
& 
\begin{tabular}[c]{@{}c@{}}
1.43 \\ 
{}[1.40--1.47]
\end{tabular}
\\
\addlinespace

Temperature ($^\circ$C)
& 
\begin{tabular}[c]{@{}c@{}}
0.54 \\
{}[0.51--0.56]
\end{tabular}
& 
\begin{tabular}[c]{@{}c@{}}
0.32 \\ 
{}[0.31--0.33]
\end{tabular}
& 
\begin{tabular}[c]{@{}c@{}}
0.54 \\ 
{}[0.52--0.57]
\end{tabular}
& 
\begin{tabular}[c]{@{}c@{}}
0.39 \\ 
{}[0.38--0.40] 
\end{tabular}
\\
\addlinespace

\bottomrule
\end{tabular}
\end{table}

\subsubsection{Downstream Tasks}

Zero-shot results for mortality risk prediction, per risk category, are shown in Supplementary Figure \ref{fig:stratified_mort}. AUROC and AUPRC results for PedsFM representation learning evaluation are shown in Supplementary Figure \ref{fig:rep_results}.

\begin{figure}[H]
    \centering
    \includegraphics[width=0.3\linewidth]{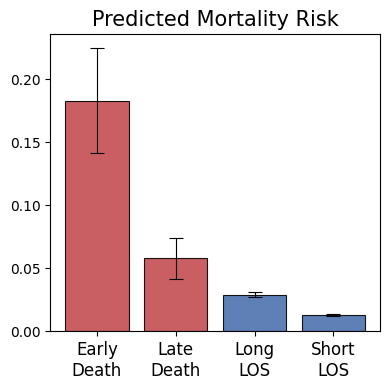}
    \caption{Stratified zero-shot risk prediction results. Early death is defined as in-hospital mortality within the first 7 days of admission, late death as in-hospital mortality after the first 7 days of admission. Similarly, long length-of-stay (LOS) is defined as stays longer than 7 days, and short LOS as stays of 7 days or less.}
    \label{fig:stratified_mort}
\end{figure}

\begin{figure}[H]
    \centering
    \includegraphics[width=0.8\linewidth]{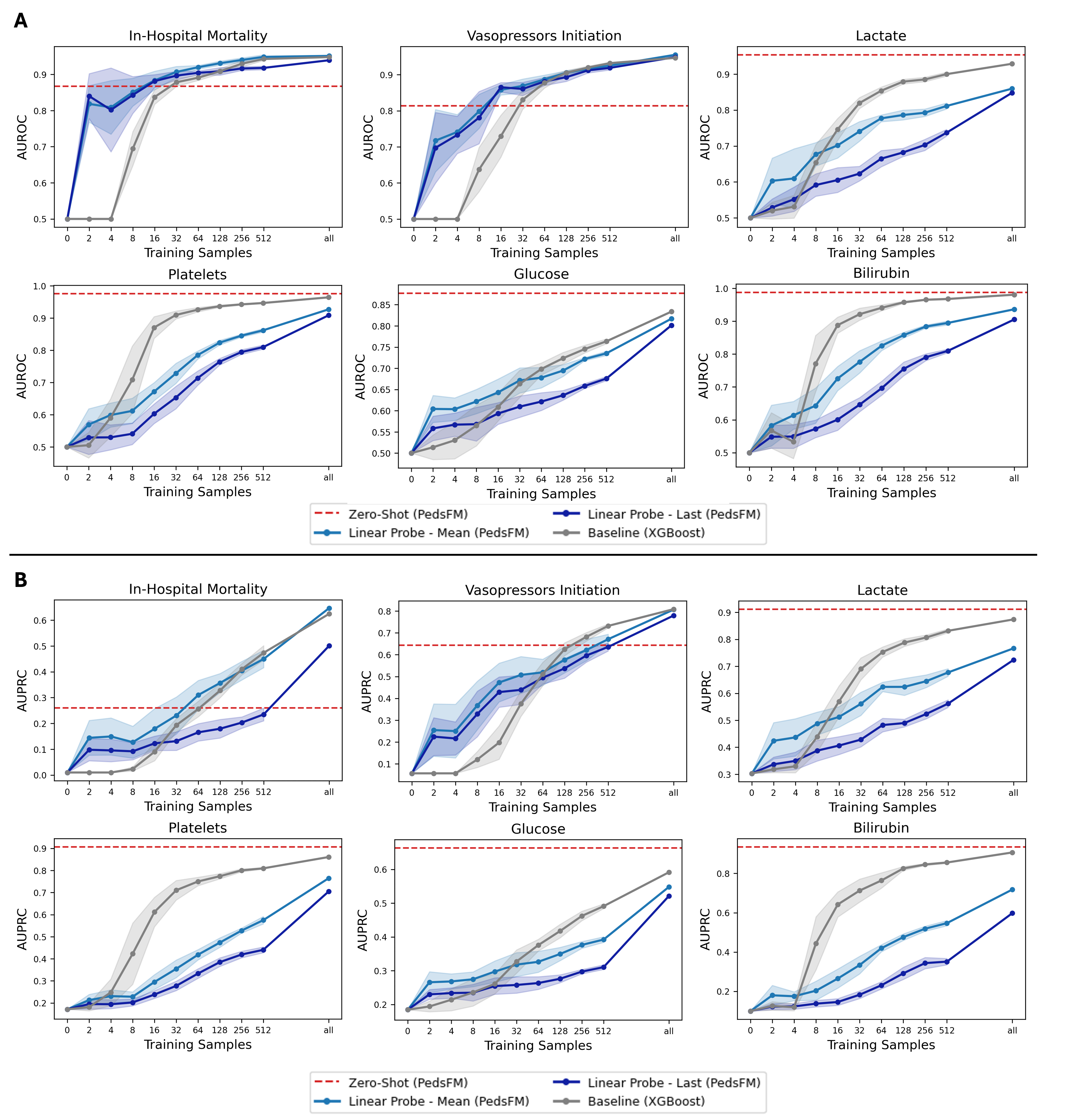}
    \caption{Representation learning results for downstream tasks across different sample sizes. A) AUROC results, B) AUPRC results.}
    \label{fig:rep_results}
\end{figure}

\end{document}